\documentclass[journal]{IEEEtran}

\ifCLASSINFOpdf
\else
\fi

\usepackage[table]{xcolor}
\usepackage{graphicx,amsmath,amssymb,epstopdf,mathtools,multirow,booktabs}
\usepackage{subcaption}	
\usepackage{amsthm,float}
\usepackage{threeparttable, tablefootnote}
\usepackage{algpseudocode}
\usepackage{arydshln}
\usepackage{pdfrender}
\usepackage{gensymb}
\usepackage[nolist]{acronym} 

\usepackage[hidelinks]{hyperref}
\usepackage{paralist}

\newacro{hbsf}[HBSF]{hybrid bending soft finger}
\newacro{spa}[SPA]{soft pneumatic actuator}
\newacro{pneunets}[Pneu-nets]{pneumatic networks}
\newacro{fea}[FEA]{fluidic elastomer actuator}
\newacro{fem}[FEM]{finite element modeling}
\newacro{3d}[3D]{three-dimensional}
\newacro{sma}[SMA]{shape memory alloy}
\newacro{pwm}[PWM]{pulse-width modulation}

\makeatletter
\def\endthebibliography{%
	\def\@noitemerr{\@latex@warning{Empty `thebibliography' environment}}%
	\endlist
}
\makeatother

\begin{document}
%
\title{A Novel Design of Soft Robotic Hand with a Human-inspired Soft Palm for Dexterous Grasping
}
%
\author{Haihang~Wang,
        Fares~J.~Abu-Dakka$^*$,
        Tran~Nguyen~Le,
        Ville~Kyrki,
        and~He~Xu

\thanks{H. Wang is  with College of Mechanical and Electrical Engineering, Harbin Engineering University, China and with Intelligent Robotics Group at the Department of Electrical Engineering and Automation, Aalto University, Finland (e-mail: wanghaihang@hrbeu.edu.cn).}

\thanks{F.J. Abu-Dakka, T.~Nguyen~Le and V. Kyrki are with Intelligent Robotics Group at the Department of Electrical Engineering and Automation, Aalto University, Finland (e-mail:  \{firstname.lastname\}{@}aalto.fi).}

\thanks{H. Xu is with Electrical Engineering, Harbin Engineering University, China (e-mail: railway\_dragon@sohu.com).}
}

%

\maketitle

\begin{abstract}
Soft robotic hands and grippers are increasingly attracting  attention as a robotic end-effector. Compared with rigid counterparts, they are safer for human-robot and environment-robot interactions, easier to control, lower cost and weight, and more compliant. Current soft robotic hands  have mostly focused on the soft fingers and bending actuators. However, the palm is also essential part for grasping. In this work, we propose a novel design of soft humanoid hand with pneumatic soft fingers and soft palm. The hand is inexpensive to fabricate. The configuration of the soft palm is based on modular design which can be easily applied into actuating all kinds of soft fingers before. The splaying of the fingers, bending of the whole palm, abduction and adduction of the thumb are implemented by the soft palm. Moreover, we present a new design of soft finger, called \ac{hbsf}. It can both bend in the grasping axis and deflect 
in the side-to-side axis as human-like motion. The functions of the \ac{hbsf} and soft palm were simulated by SOFA framework. And their performance was tested in experiments. The 6 fingers with 1 to 11 segments were tested and analyzed. The versatility of the soft hand is evaluated and testified by the grasping experiments in real scenario according to Feix taxonomy. And the results present the diversity of grasps and show promise for grasping a variety of objects with different shapes and weights.
\end{abstract}


\IEEEpeerreviewmaketitle

\section{Overview and Motivation}
%
%
%
%
\IEEEPARstart{H}{ands} or grippers are essential component for robotic manipulation as they handle objects with certain positions, orientations and contact forces. They serve as the end-effector interfacing between target objects and robots. Dexterous grasping is a prerequisite for task-dependent manipulation, which requires the consideration of important factors such as the interaction-force, stiffiness/compliance, dexterity and number of degrees of freedom \cite{Raphael35}. 

\begin{figure}[t]
    \centering
    \includegraphics[width=.8\linewidth]{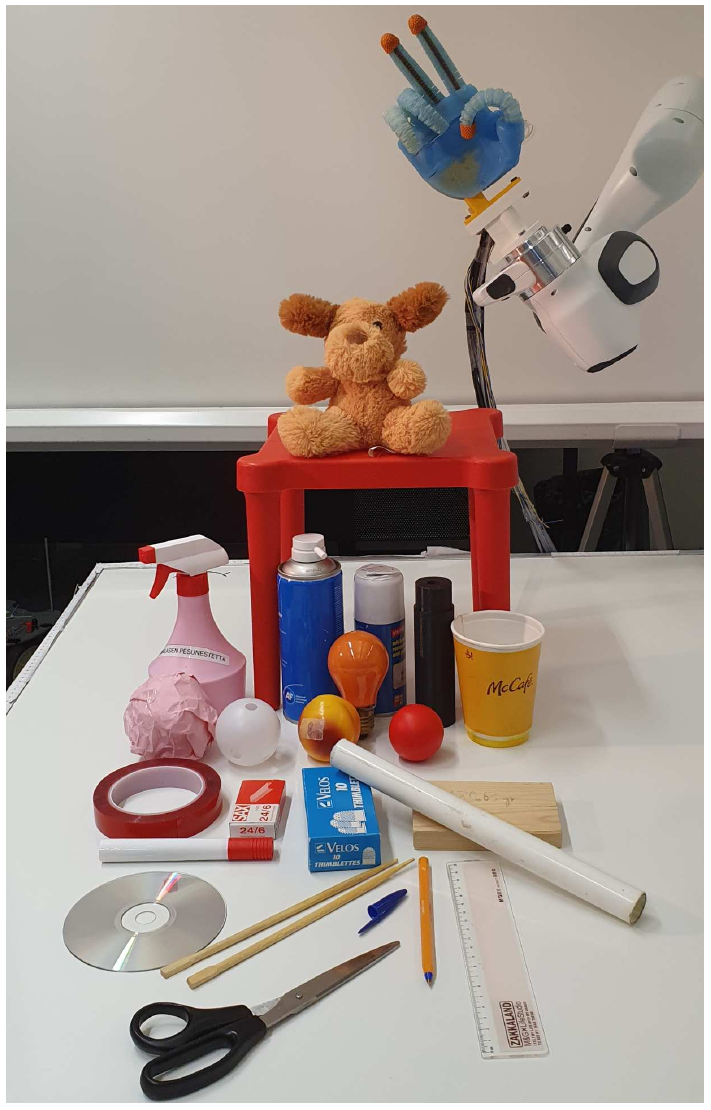}
    \caption{The soft humanoid hand and the grasping objects used in this work.}
    \vspace{-.5cm}
    \label{fig:handandobjects}
\end{figure}

Conventional rigid robotic hands for industrial applications are generally able to provide high accuracy in position thanks to their sophisticated actuation and sensing mechanisms. However, it is hard to control the contact force between the rigid hand and objects as the rigid structure driven by electrical motors commonly generates large contact forces. In real-world scenarios, usually, we require grippers to manipulate objects with uncertain shapes, sizes and poses in uncertain environments \cite{Xiaomin1}. Moreover, when the targeted objects are fragile or delicate, large contact forces can deform or even damage the objects. Another drawbacks of these rigid hands lie in their heavy weight and high cost. Thus, the applications of soft robotic hands with passive compliance have attracted attention for an inherently safe and adaptive contact. Soft robotic hands not only can easily adapt to objects of various shapes and sizes, but also can perform a self-adaptive contact without the need of sophisticated control as rigid hands. Furthermore, their soft nature helps to minimize the damage to the manipulated objects.
 
\begin{figure*}[t]
    \centering
    \includegraphics[width=1\linewidth]{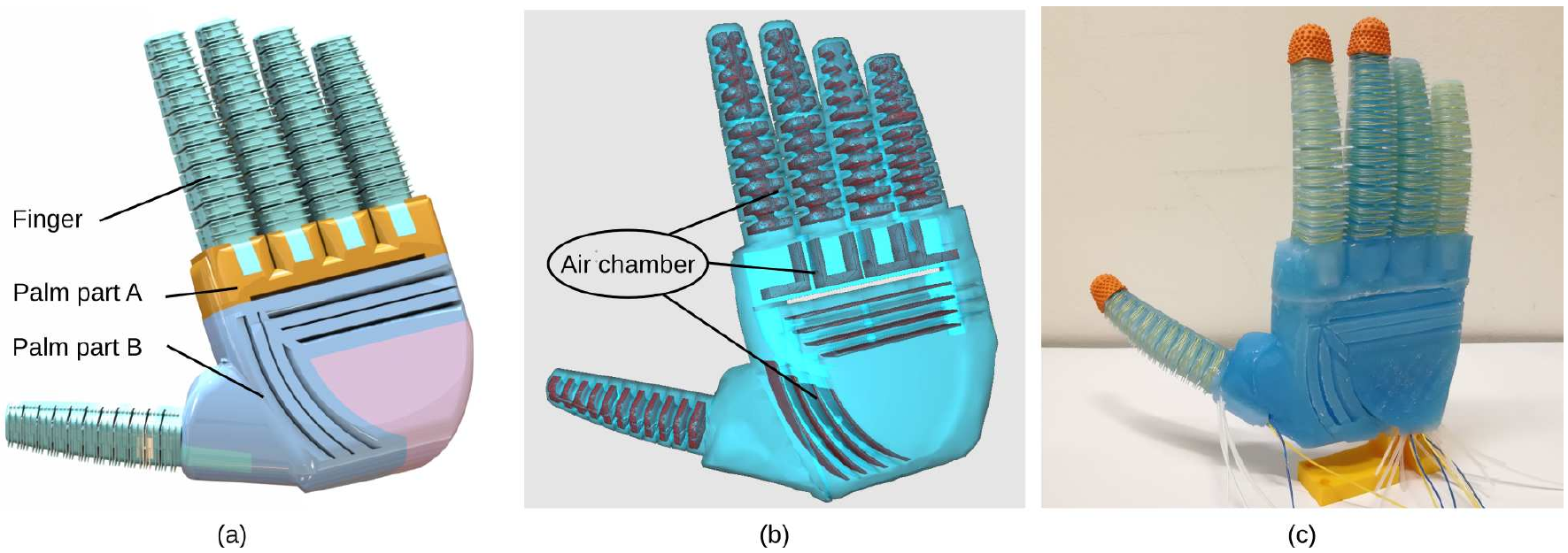}
    \caption{The proposed pneumatic soft humanoid hand with dexterous palm. (a) Illustrates the 3D model of our hand.
    (b) Shows air chambers: the red parts show the inner distribution of the air chambers, which will be pressurized based on SoftRobots plugin. (c) The soft robotic hand prototype.}
    \vspace{-.5cm}
    \label{fig:overallStructure}
\end{figure*}

Soft bending actuators, used as fingers, are the main component of soft robotic hands/grippers. They can be sorted as different types, such as \acp{fea}, cable-driven actuators, \acp{sma}, electromagnetic/magnetic actuator \cite{Shintake30}. Amongst these, \acp{fea} have achieved particular sizable push toward the utilization of compliant hands. \acp{fea} are mainly made of silicone rubber and driven by pneumatic or hydraulic. The pneumatic type of \acp{fea} are known as \acp{spa}. The most popular \acp{spa} are the \ac{pneunets} bending actuators designed by Whitesides et al. \cite{Mosadegh24} and fibre-reinforced actuators designed by Galloway et al. \cite{6766586}. \ac{pneunets} is bonded by 2 layers: the silicone-based top layer containing numerous chambers inside (like networks) and the inextensible bottom layer. When the actuator is inflated, the top layer will extend, and the actuator will achieve a bending motion. Fibre-reinforced actuator comprises an extensible chamber, an inextensible layer and fibres. Its bending mechanism is similar to the \ac{pneunets} actuator. 
The fibre-reinforcement is used to limit the chamber in axial extension instead of useless radial expansion. Both types are simple in design, effective and easy to fabricate. 

In literature, there are different design and application for soft robotic hands based on Pneu-nets bending actuator \cite{Lotfiani7, Abondance855,Yilin97,Clark65} and fibre-reinforced actuator prototypes \cite{Raphael35,  Xiaomin1, Polygerinos94, Fras03, Mahmoud77,Galloway19,Weiping015}. However, compared with each other, \ac{pneunets} actuator has a lower capacity of input pressure due to its total soft top layer under the same wall thickness, which limits its maximal grasping force. 
In addition, fibre-reinforced actuator has a lower bending efficiency, which limit its bending angle under the same pressure. In order to overcome the
shortcomings of them, in this paper, we design a novel \acf{hbsf} by  integrating the inner chambers network structure inspired by \ac{pneunets} with fibre-reinforcement method.

Recently, there have been rapid developments on soft robotic hands and grippers. However, most of them focus on the study of soft fingers and overlook the importance of the palm. The fingers were usually assembled together and fixed in a rigid palm or basement. However, the palm plays a considerable role in the grasping functioning. The fixed position of fingers will greatly limits the grasping scope and pose of the hands.

In order to achieve a dexterous manipulation, we refer to the postural variability of the hand: the higher this variability, the more dexterous we consider a hand (for example of grasping postures referring to the grasp taxonomies purposed by Feix et al. \cite{Feix27}). 
Robotic hand with a changeable palm can adjust the position and orientation of fingers, which can significantly improve the postural flexibility of a robotic hand in terms of sizes and shapes of objects which can be grasped.  
Sun et al. \cite{Yilin97} presented a flexible robotic gripper with a rigid changeable palm. The distance of the fingers can adjust using slider and beam mechanism. An opposable thumb is important and useful to achieve dexterity in robotic hand. However, the rigid changeable palm can only change the position of the fingers.
RBO hand 2 \cite{Raphael35} has a soft palmar actuator for enabling thumb abduction. They proposed a new PneuFlex actuator with fiber-reinforcement as the soft fingers. In addition, they also use two connected PneuFlex actuators as the base of the thumb to achieve the dexterity of the thumb. The other four fingers were fixed on a 3D-printed scaffold. The assembly  angle between thumb and the other four fingers is about 120\degree, instead of a plane.
The soft biomimetic prosthetic hand developed by Fras et al. \cite{Fras03} presents a similar design of the thumb abduction. They used the PneuFlex actuators as in \cite{Raphael35}. They applied one  for the thumb and two for the palm. The exoskeleton for fixing the soft actuators is deformable and based on a 3D scan of a real human hand. Both hands have soft actuators for thumb abduction. However, the palm motion for the other four fingers were ignored. The four fingers of their hands can only bend to a certain direction. 
In contrast of \cite{Raphael35} and \cite{Fras03}, in order to augment the human-like palmar function of the soft hand, we present a pneumatic soft humanoid palm that can help thumb abduction, the four other fingers splay and palm bend.

In this paper, we propose a compact hybrid solution of soft humanoid hand, as shown in Figure 
    \ref{fig:handandobjects}. For the sake of clarity, the main contributions of this paper are:
\begin{itemize}
    \item [--] A novel design of \acf{hbsf} by  integrating the inner chambers network structure with fibre-reinforcement method.
    \item [--] A novel design of pneumatic soft humanoid palm.
\end{itemize}
The soft robotic hand is made of soft materials only.
The \ac{hbsf} can robustly grasp a variety of objects with different weights, sizes, shapes and stiffness. 
The soft humanoid hand consists of 5 soft pneumatic fingers and 2 parts of the soft palm, which are all independent and assembled together using silicone connections.

\begin{figure*}[htbp]
    \centering
    \includegraphics[width=.9\linewidth]{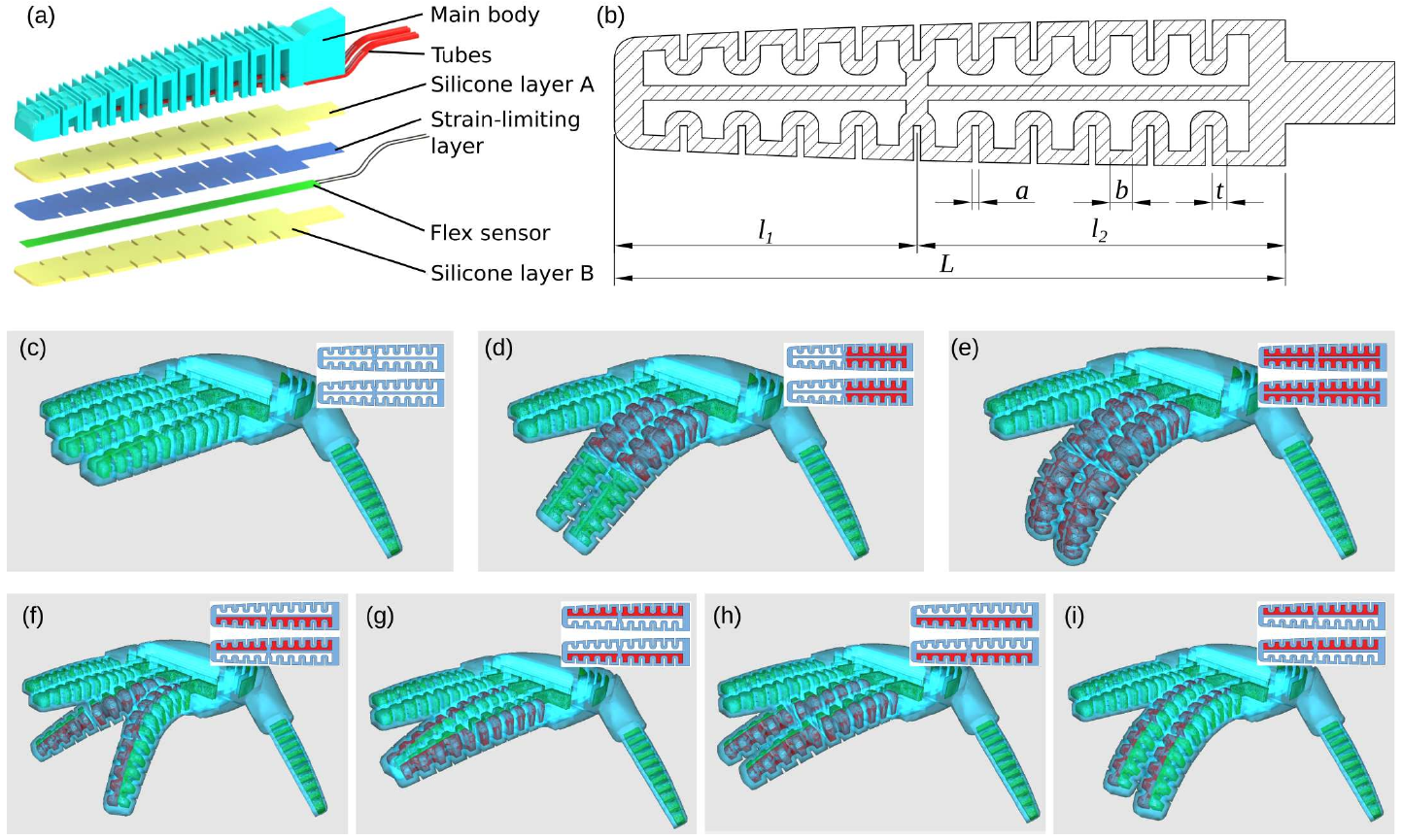}
    \caption{The structure of the \ac{hbsf}. (a) The schematic illustration of the components of \ac{hbsf}. The fibre-reinforcement structure is presented by the thin raised features on the main body's surface. (b) The sectional drawing (bottom view). The red chambers in (c)-(i) present the pressurized actuators in SOFA \cite{allard2007sofa}. (c) The original morphology.  (d) Pressurize the 2nd section of the index and middle fingers. (e) Pressurize the two sections together. (f) Deflection of two fingers to enlarge the grasping region.  (g) Deflection of two fingers to decrease the gap between them. (h)-(i) Deflection of two fingers to achieve their wiggling to the left and to the right.
    }
    \vspace{-.5cm}
    \label{fig:designoffinger}
\end{figure*}
\section{Design and Simulation}
\label{sec:design}

\subsection{Soft hand}
\label{sec:structure}
The design of soft hands/grippers can be divided into two main morphological types: the anthropomorphic hands and the grippers with several spatially evenly distributed fingers. For our hand, we choose an anthropomorphic design in shape with a new dexterous soft robotic palm. Figure \ref{fig:overallStructure}(c) presents the prototype of the soft humanoid hand, which has a weight of only 300 g and is about 1.2 times the size of a typical human hand. 

As shown in Figure \ref{fig:overallStructure}(a), the soft humanoid hand in this paper has three functional components: finger, palm part A and palm part B. All these components are actuated pneumatically. The fingers are designed to grasp, grip and manipulate targeted objects through bending in the grasping axis and deflecting in the side-to-side axis. The five fingers of the hand are sorted as thumb and four planar fingers (index, middle, ring and little finger). Palm part A is used to splay the four planar fingers, which 
extend the distance between fingers and enlarge the grasping scope. Palm part B achieves two functions: palm bending and thumb abduction and adduction.

Simulation was conducted to guide the design and to ensure the proposed design can work as intended in real-time. We used \ac{fem} to simulate and analyze the whole soft hand by following the analysis in \cite{Duriez138}. The simulation in real-time was implemented in SOFA framework \cite{allard2007sofa} with SoftRobots plugin \cite{Coevoet362}. The mesh file of the soft hand consists of 40316 tetrahedra and 11894 nodes.

\subsection{Soft finger}
\label{subsec:finger}

\subsubsection{Function}
In this work, we aim to develop a humanoid pneumatic soft hand with dexterous soft fingers. Among the five fingers, the functions of the thumb, index and middle fingers are the main ones for grasp manipulation, while ring and little finger are usually play an assistant role. 
Index and middle fingers have two functional goals: bending in the primary grasping axis (toward  the soft palm) and deflecting from side to side (perpendicular to the primary grasping direction). The function of the ring and little fingers in this work is only bending motion.

Bending motion is the basic and main function for grasping. The bending angles are set to about 180\degree like human fingers. Side-to-side deflection motion is also useful and necessary to preform splaying and gripping between two adjacent normal planar fingers and rotatory movement of the object among three fingers' grasping.

\subsubsection{Design}
The soft hand fingers, which are able to both bend and deflect, are made of \ac{hbsf} based on a modular approach. \acp{hbsf} are used for the thumb, index and middle fingers. The structure of the 3 fingers is  identical in shape with a length $L$ of 90 mm, while the thumb is 20 mm shorter than the index and middle fingers. The structure of ring and little fingers is a simplified version of \ac{hbsf}, which can only bend to the main direction. The two side by side chambers in one segment of \ac{hbsf} was omitted into one chamber.
 
In this work, the \ac{hbsf} combines the advantages of \ac{pneunets} bending actuators \cite{Mosadegh24} and fibre-reinforced actuators \cite{6766586}. About the configuration, we refer to the humanoid morphology of the soft fingers in Deimel et al. \cite{Raphael35}. 
As shown in Figure \ref{fig:designoffinger}(a), the main body of \ac{hbsf} consists of silicone materials, a strain- limiting layer in inextensible but flexible materials (silk screen) in its bottom, and the fibre-reinforcement thread. The strain- limiting layer determines the bending motion of the finger. And the fibre-reinforcements can protect the silicone-based chambers from excessive  expansion.

The tubes are used to pressurize the air cavities inside fingers. A flex sensor (Spectra Symbol FS-L-0095-103-ST) is glued to the bottom flat face. The structure of the \ac{hbsf} is like a bellow.  From the sectional drawing in the bottom plane in Figure \ref{fig:designoffinger}(b), the most recent version used in index/middle fingers has 11 short bellows (called as 11 segments in this paper) with a wall thickness $t=2$ mm. The two sections of the finger are shared with the same length, which means $l_1=l_2$. The little gap between segments $a$ is equal to 1 mm, which is limited by the thin-wall strength of the 3D printed molds. The distance between each short bellows $b$ is the key parameter, which will affect the bending and deflection performance of the \ac{hbsf}. A \ac{hbsf} is divided into four air cavities.

\subsubsection{Simulation}
Figure \ref{fig:designoffinger}(c)-(i) present the simulation results of the soft fingers. Because the SOFA framework is developed for simulation in real-time, the fibre-reinforcement structure is so thin to mesh it into element based on the \ac{fem}. Thus, it is omitted in the simulation. As a result, the input air pressure in SOFA simulation cannot fit the capability of the \ac{hbsf} in real-world, which limits the bending angle of the fingers in the simulation results. 

The fingers' deflection enables several useful collaborative motions between the index and middle fingers. Although the motions in Figure \ref{fig:designoffinger}(f) and (g) cannot work as a gripper due to the limitation of the low stiffness of the soft fingers in the side-to-side axis, they play a key role in achieving the Feix \cite{Feix27} grasp postures 14, 17, 20, 21 and 23  in Figure \ref{fig:experimets photos}.

In addition, the deflection of the thumb acts  similar to the functions shown in Figure \ref{fig:designoffinger}(h) and (i). The thumb deflects toward the palm for grasping small objects, while deflects away from the palm for grasping bigger objects.

\subsection{Soft palm}
\label{subsec:palm}
\subsubsection{Function}
For human hand grasping, the palm always collaboratively works with the fingers to implement all kinds of manipulation. In this paper, the functions of the soft palm is inspired from human palm motion, which include three main functions: thumb abduction, four planar fingers splaying and palm bending. 
A key feature of the human hand is the thumb abduction and adduction. This feature allow the thumb to rotate and has a proper position and orientation with respect to the manipulated object, which is essential of grasping.

The splaying of the four fingers enlarges the capability of grasping different size objects. By increasing the angle between the adjacent fingers, the contacting force from the four planar fingers  will evenly distributed on the objects and it also make it possible to grip larger objects using two adjacent fingers. 
The bending of the whole plane of the palm is beneficial to facilitate the grasping of smaller objects. Without the help of palm bending, the thumb of soft hand cannot touch the other fingers even all the fingers are pressurized under the highest pressure and bending with 180\degree. 

\subsubsection{Design}
\label{subsubsec:Design of soft palm}
\begin{figure}[tbp]
    \centering
    \includegraphics[width=.9\linewidth]{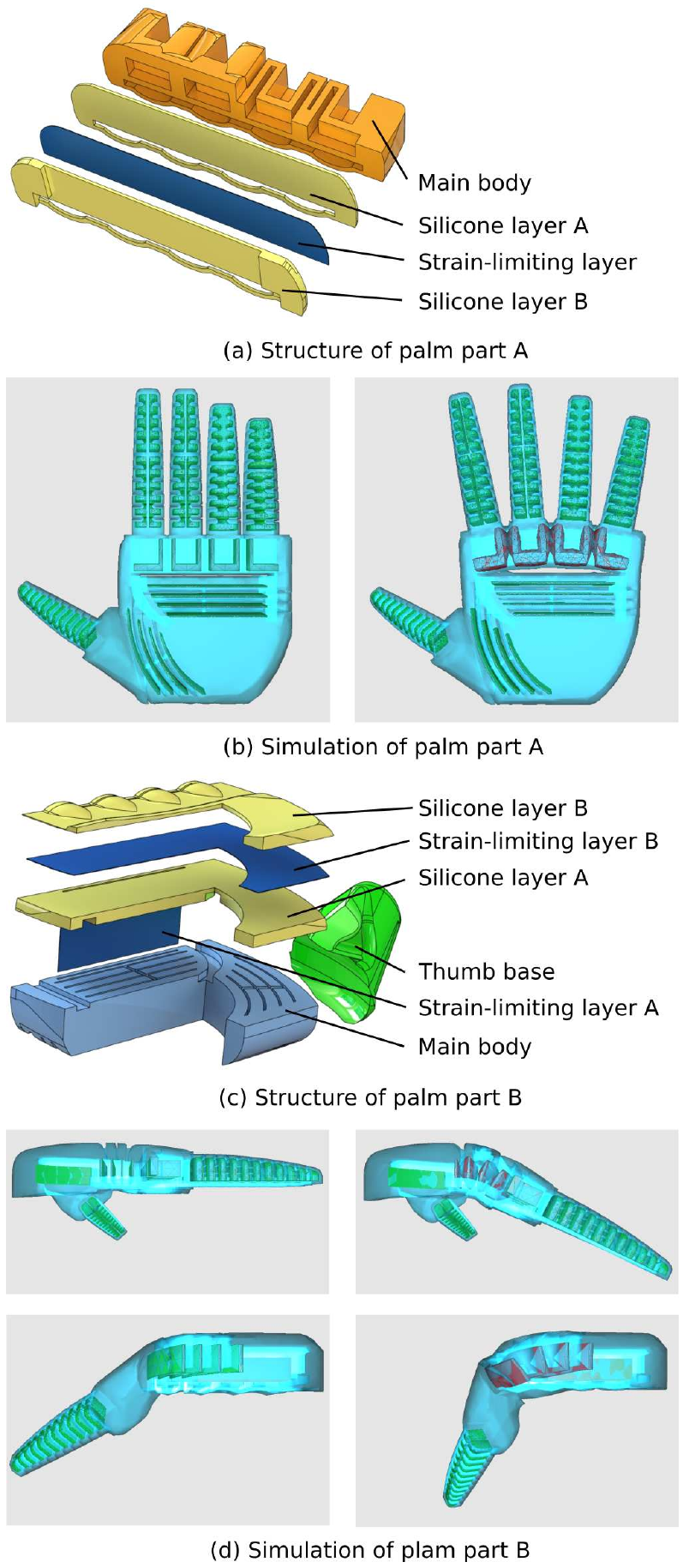}
    \caption{The structure and simulation of the soft palm. (b) and (d) Show a comparison of the palm actuators under inflation and deflation states. The red chambers in (b) and (d) presents the pressurized actuators in SOFA.}
    \vspace{-.5cm}
    \label{fig:design of soft palm}
\end{figure}

The method of modular design are applied in the design of soft palm. The three functional goals are divided and embedded into two parts: palm part A and B, as shown in Figure.  \ref{fig:design of soft palm}. The structural design of both two parts are inspired by the mechanisms of \ac{pneunets} actuators, consisting of networks-like chambers inside the silicone body and the strain-limiting layer on the bottom. 

\begin{figure*}[htbp]
	\centering
	\includegraphics[width=1\linewidth]{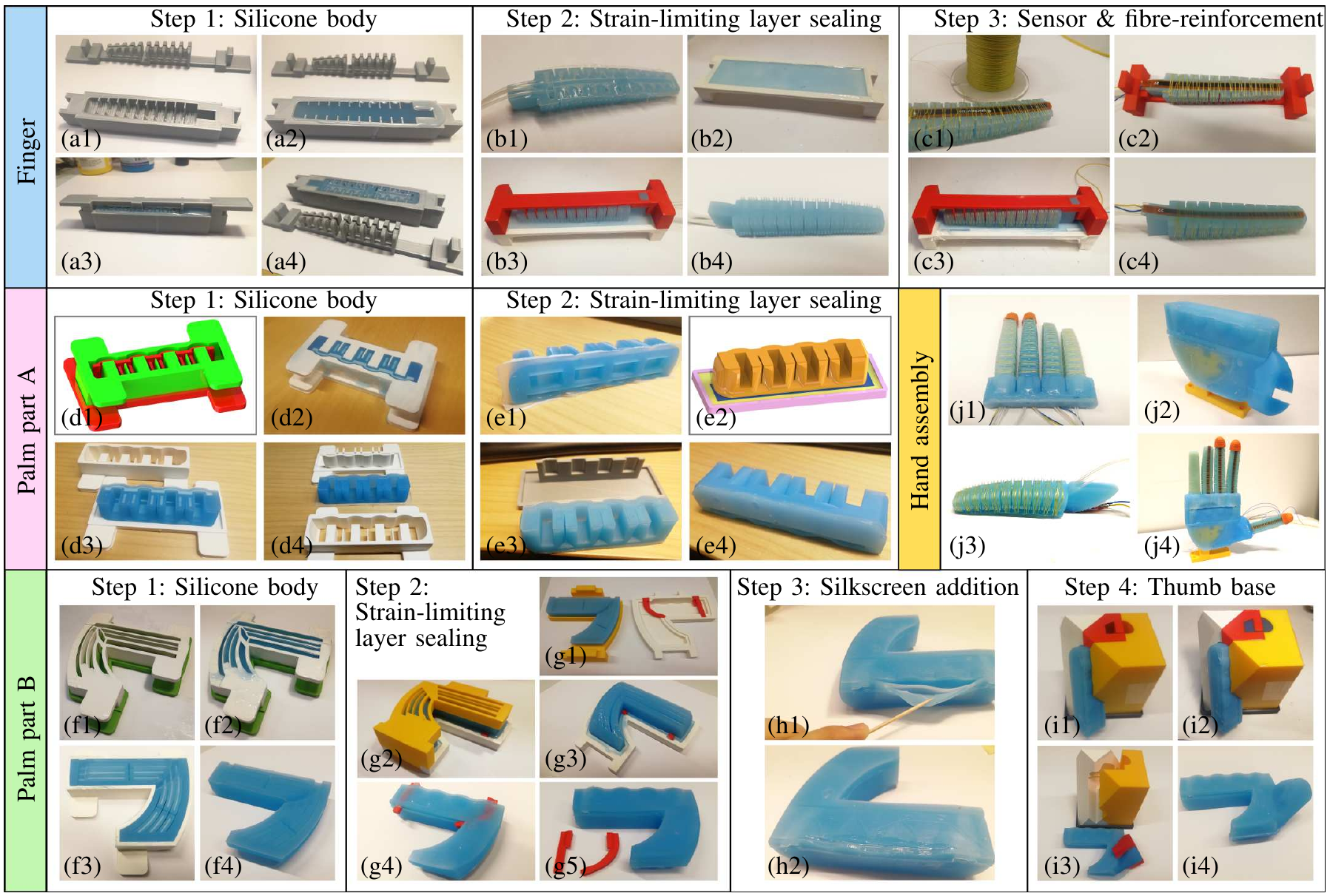}
	\caption{The fabrication process of the soft hand. \emph{Finger}: (a1)-(a4) show that the main body of the \acp{hbsf}. It is made by first pouring silicone material in its bottom mold and then press the top mold. In Step 2, the red part in (b3) is for clamping and positioning the main body. (c1)-(c4) is to fix the fibre and flex sensor  to prevent relative displacement when the finger bends.
	\emph{Palm part A}: (d1) The molds. (d2) Pure silicone material. (d3)-(d4) Unmold. In Step 2, the 3D printer support is used to hold and ensure the thickness of the bottom surface.
	\emph{Palm part B}: The process of making the main and bottom body is similar with the other parts. The 2 red sticks in (g1)-(g5) are to reserve the space for air tubes and electric wire. The yellow support in (g2) works as a similar role of the red in (b3).
	\emph{Hand assembly}: (j1) The assembly of palm and 4 planar fingers. (j4) The final result.
	}
	\vspace{-.5cm}
	\label{fig:fabricaiton}
\end{figure*}

Palm part A achieves the splaying of the four planar fingers, while part B implements the abduction of the thumb and overall bending of the palm. 
Instead of bending as \ac{pneunets} actuators, the palm part A splays the four planar fingers with a small hump after pressurized. As shown in Figure  \ref{fig:design of soft palm}(a), the main body has four air chambers to be actuated. The four large groove in front are used to fix the four fingers. The bottom cavity is designed to arrange the pneumatic tubes and sensor wires inside the hands.  
As for part B, the bending motion of the palm is achieved by enlarging the horizontal width of the \ac{pneunets} actuator. The width of the air chambers along the palm were designed as wide as possible to generate enough force to bend the  palm with four planar fingers. The tubing tunnel connecting with the bottom cavity in part A is used to run the tubes and lines through the palm.

The soft palm uses the same key geometric parameters with the soft fingers, such as the wall thickness $t$ and the little gap between segments $a$. The stiffness of the palm for enabling a reliable support for the other fingers is considered and simulated in SOFA framework. In addition, the humanoid appearance design were implemented at last, after ensuring the soft actuators work as intended and it will not affect the performance of the pneumatic actuators. 

\subsubsection{Simulation}
Figure  \ref{fig:design of soft palm}(b) and (d) present the simulation results of the two parts of the soft palm. The three functions of the palm can be clearly detected through the comparison of each actuators before and after being pressurized.

\section{Fabrication}

\subsection{Actuator body molding}
The soft body was fabricated using the silicone rubber (Dragon Skin 10 Medium) by Smooth-On (Sil-Poxy) with a 10 Shore-A hardness. Parts (A and B) were mixed (1:1 ratio) in a plastic cup and then put into a vacuum chamber. Using vacuum pump, the air trapped in mixed silicone materials will expand to bubble and finally collapse under about 0.9 bar vacuum pressure. Afterward, the silicone material was poured into the molds as described in Step 1 and 2 of finger and palm parts in Figure \ref{fig:fabricaiton}. The molds are generally left in an upright position to cure for five hours at room temperature.

\subsection{Strain-limiting layer sealing}
The strain-limiting layer is used to ensure the actuators bending in the desired direction during pneumatic pressurization \cite{Mosadegh24}. Its material is silkscreen fabric. It is placed in the bottom mold of the Step 2 in Figure \ref{fig:fabricaiton} and glued together with the bottom sealing portions of all three components. Subsequently, the fabrication of the main body of the actuators is finished and the chambers inside the silicone body are all sealed. It is noted that the Step 3 of Palm part B does not need molds. As shown in in Figure \ref{fig:fabricaiton}(h1)-(h2), the silkscreen fabric  is added manually in the reserved gap. Then, after pouring silicone into the gap and curing, the silkscreen fabric is combined with the main body.

\subsection{Assembly}
Figure \ref{fig:fabricaiton}(d1)-(d4) shows the main procedure of the hand assembly. There are five soft fingers and two soft palm parts in this hand. We used the same silicone materials to  bond different components together by manually pouring liquid silicone on their connecting surface. As shown in Figure \ref{fig:fabricaiton}(d1), the end of four planar fingers is fixed on the large groove mentioned in Section \ref{subsubsec:Design of soft palm}. The tubes of each finger are arranged into the bottom cavity inside the palm part A.

Figure \ref{fig:fabricaiton}(d2) presents the assembly result of the palm part B and a soft-rigid hybrid support. The support consists of silicone skin fitting with the shape of the palm part B and the rigid 3D-printed skeleton inside. All the pneumatic tubes will go through the support and run out the hand together. The yellow 3D-printed skeleton of the support is used as the base to fix the hand with the robot arm. As a result, the three parts in Figure \ref{fig:fabricaiton}(d1)-(d3) are assembled together into a soft humanoid hand, as shown in Figure \ref{fig:fabricaiton}(d4).

\section{Control}
\label{sec:control}

\begin{figure}[!htbp]
    \centering
    \includegraphics[width=.8\linewidth]{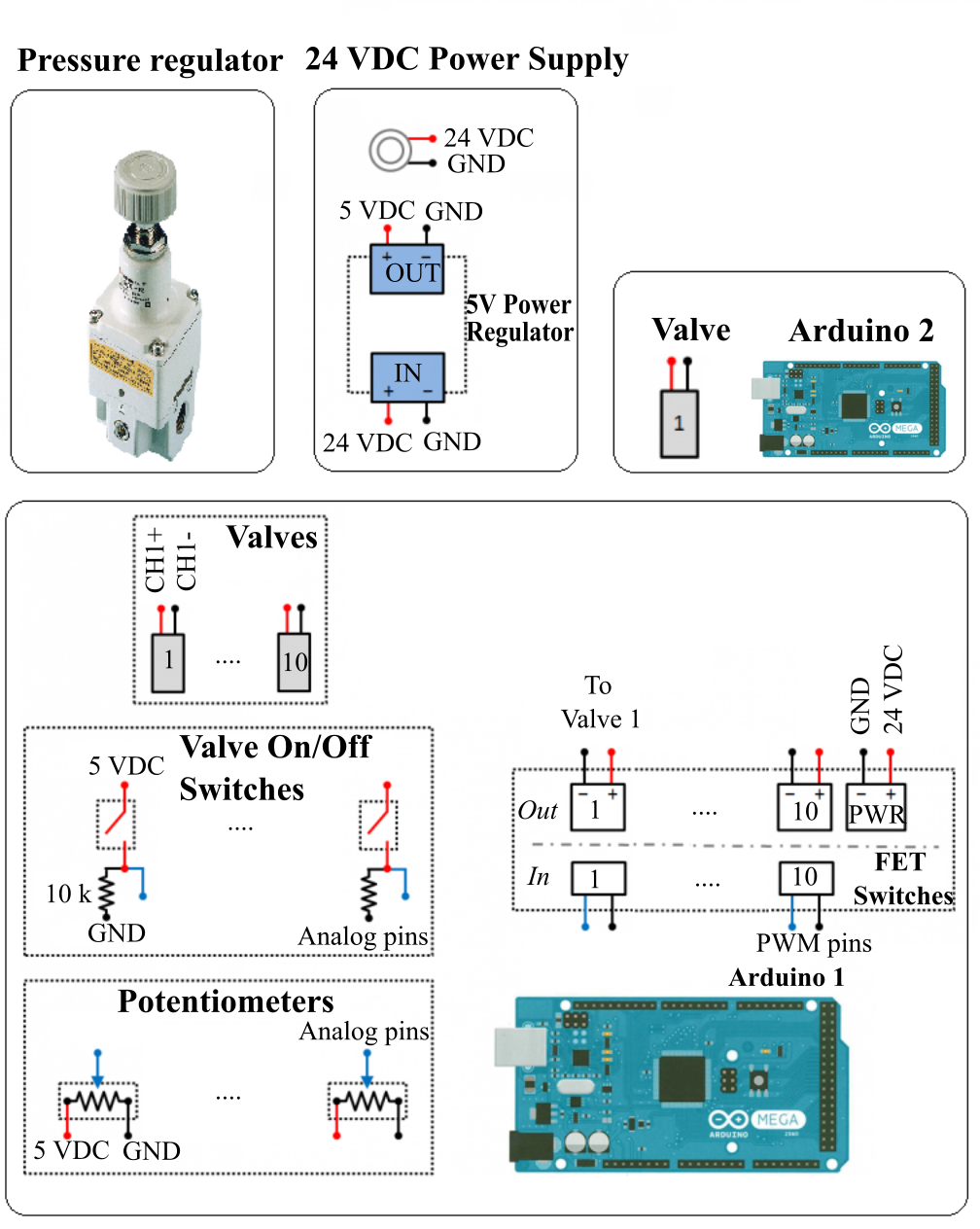}
    \caption{The architecture of the implemented controller board. For clarity, we only show the connection scheme of the Arduino 1. The connection scheme of the Arduino 2 is identical to that of the Arduino 1.}
    \vspace{-.5cm}
    \label{fig:controlboard}
\end{figure}

To control the proposed soft hand at different operating conditions, a
controller platform is constructed. As the designed soft actuators are pneumatic, controlling the pressure and the duration of the input pneumatic supply
is needed. The controller platform is implemented based on the proposed design of the soft robotics toolkit\footnote{Fluidic Control Board, \url{https://softroboticstoolkit.com/book/control-board}}. The architecture of our controller board is shown in Figure \ref{fig:controlboard}. 

The controller board consists of a pneumatic regulator (which regulates the pressurized air to the system), a set of solenoid valves\footnote{\raggedright SMC-VQ110U-5M Solenoid valve, \url{https://www.smcpneumatics.com/VQ110U-5M.html}} (which can open and close to direct the flow of air into the system). The valves are powered and directed by power FET switches. As we want the hand to have as much degree of freedom as possible, we use 20 solenoid valves in total. Thus, each chamber of the finger can be pressurized independently, which in turn, allow the hand to have more postures. Two Arduinos Mega 2560 REV3 controller are used to enable users to interface with the hardware via a serial port connection. The board can be controlled manually (by adjusting switches and potentiometers) or automated via a programmed software.

The system pressure is regulated with \ac{pwm}, which basically controls the opening and closing times of the valves, at a rate of 60 Hz through the Arduino boards. \ac{pwm} can be expressed as a technique for getting analog results with digital means. One of the most important terms in \ac{pwm} is the duty cycle. The duty cycle is the proportion of 'on' time to the regular interval or 'period' of time. Duty cycle is expressed in percent, 100\% being fully on, and 0\% being fully off. By modulating the value of the duty cycle, analog values can be achieved. For example, the valve fully closes at 0\% duty cycle, fully opens at 100\% duty cycle and opens halfway at 50\% duty cycle. Thus, the fixed regulated input pressure is set to the desired value based on the duty cycle of the \ac{pwm} signal. With this technique, the finger can easily be controlled to a certain bending angle.

\section{Experiments}
\label{sec:Experiments}
\subsection{Analysis of the HBSF}
\begin{figure}[!tbp]
    \centering
    \includegraphics[width=1\linewidth]{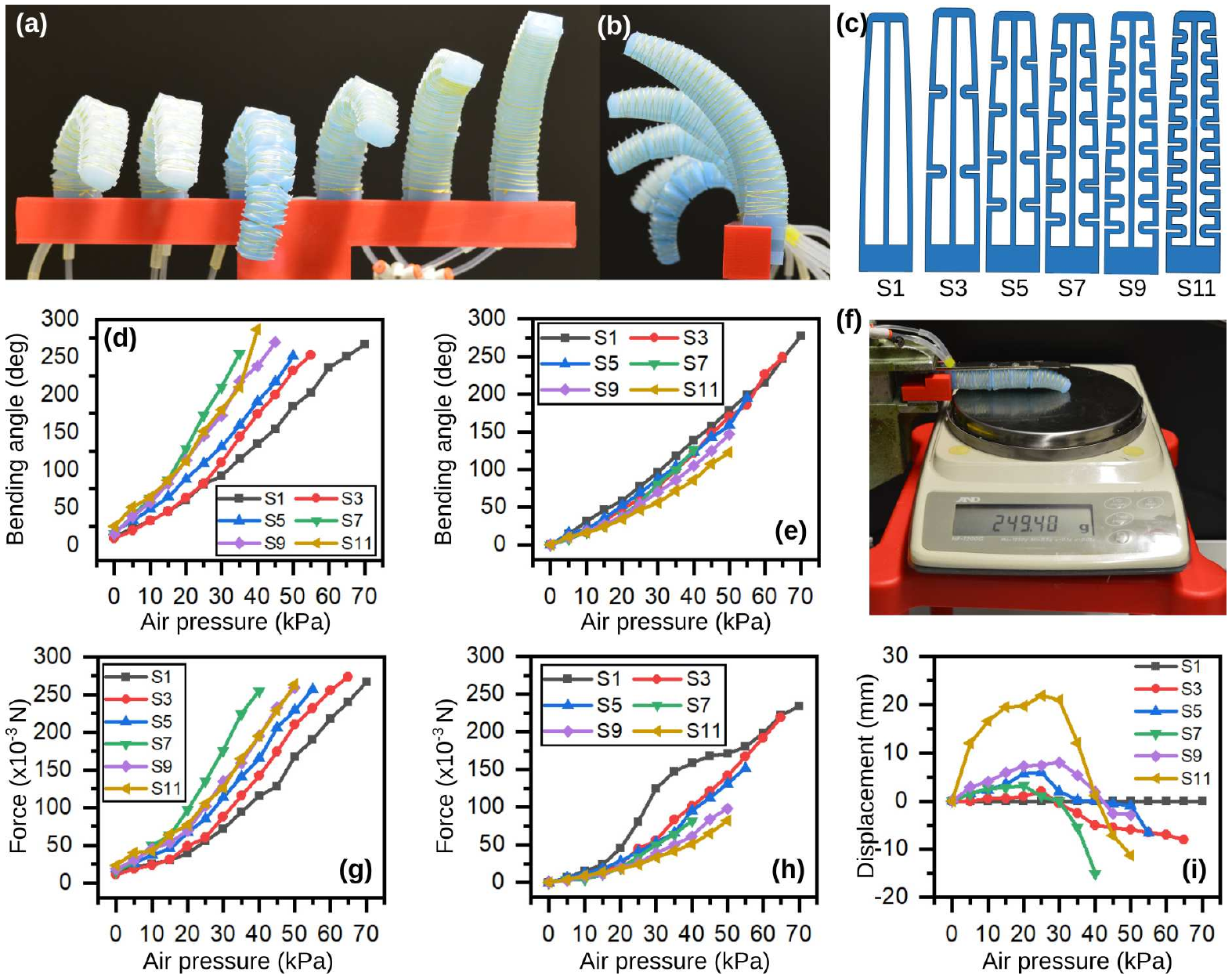}
    \caption{The analysis of the effect of the segments number on the bending and deflection motions of the six fingers. (a) The comparison of the deflection extent from front to back, respectively S11, S9, $\cdots$, S1; (b) The comparison of the bending angle from front to back, respectively S1, S3, $\cdots$, S11. (c) The sectional drawing of the six fingers. (f) The experimental setup for measuring the force of the fingers while the top layer of the fingers was constrained.  Variation of the (d) bending angle and (g) generated force of fingers while applying air pressure equally to both left and right chambers. Variation of the (e) bending angle, (h) generated force of fingers and (i) displacement of the fingertip while applying air pressure to single chamber.}
    \vspace{-.5cm}
    \label{fig:experiment of finger}
\end{figure}

The \acp{hbsf} in this paper have  novel motion. They deflect in the side-to-side axis, while the bending motion in the grasping axis is still the main function of the soft fingers. After pressurized, the adjacent inflated segments will  generate a mutual force, which achieve the deflection motion. In addition, the number of segments depends on the distance $b$, as shown in Figure \ref{fig:designoffinger}(b). When the finger length is defined, larger $b$ leads to a less segment number.  When the finger only has 1 segment, it changes
 back to the original paradigm of PneuFlex actuator with two chambers. 
In order to analyze the influence of the segments number on the deflection, six soft fingers with 1, 3, 5, 7, 9, 11 segments are fabricated and tested, as shown in Figure \ref{fig:experiment of finger}. The bending angle increases non-linearly with the applied air pressure, with a relatively slow increase in low pressure region and a rapid increase afterward when the pressure reaches beyond 20 kPa. It is obvious that the variation between the force and bending angle has a close positive correlation. However, it is also noted that under the same air pressure, the fingers with more segments generate a larger bending angle and force when pressurizing both left and right chambers but generate a lower angle and force when pressurizing single chamber. This phenomenon actually reflects the different mechanisms of the bending and deflection motion. The more segments means more grooves that mutually swell during inflation working as PneuNets actuators and less little sections  that extend during inflation working as fibre-reinforcement actuators. As shown in Figure \ref{fig:experiment of finger}(h), the generated grasping force of the S1 finger becomes larger than those with multi-segments, when the applied air pressure goes from 20 to 50 kPa.

Figure \ref{fig:experiment of finger}(i) shows the deflection displacement in the side-to-side axis. The S11 finger has significantly better performance for deflecting up to 20 mm. Combining the experimental results of bending angle and force, the S11 finger ranks the second that is easily driven under double-chamber actuation and present the minimum bending angle under single-chamber actuation. Both characteristics are conducive to enhance the deflection region of the \acp{hbsf}. It is also notable that the finger will first wiggle outwards in the initial stage of the actuation with lower air pressures and turn back when the bending angle of the \acp{hbsf} becomes large at higher air pressure, resulting in losing efficacy of the deflection motion and even get just the opposite displacement. As a conclusion, we selected the \ac{hbsf} with 11 segments in our new hand.

\subsection{Analysis of the palm}
\begin{figure}[!tbp]
    \centering
    \includegraphics[width=\linewidth]{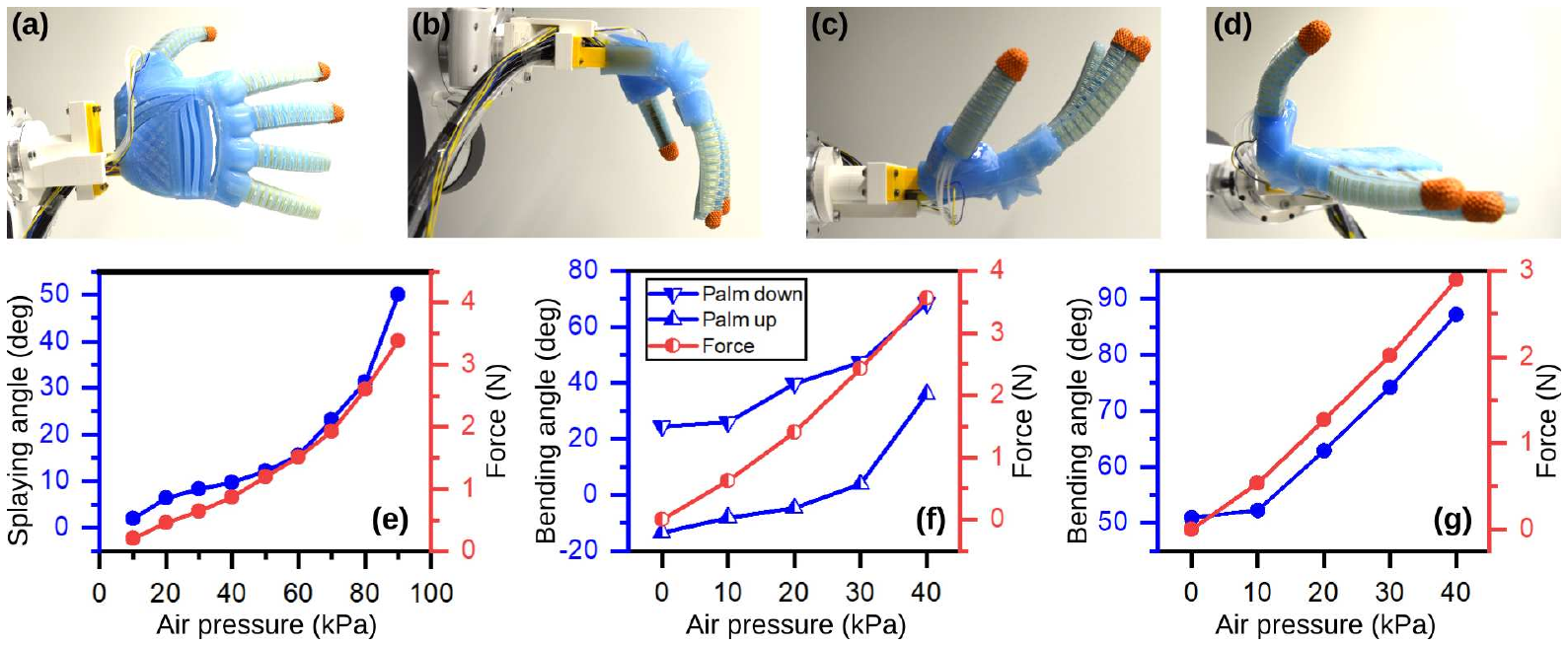}
    \caption{The validation of palm performance on bending angle and output force. (a,e) palm splaying; (b,c,f) palm bending with palm down and palm up; (d,g) thumb abduction.}
    \vspace{-.5cm}
    \label{fig:experiment of palm}
\end{figure}

Figure \ref{fig:experiment of palm} presents the experimental results of the soft palm. Different from the normal fixed palm, almost all the parts of our soft palm is deformable, which greatly increases the diversity of the hand posture.

Figure \ref{fig:experiment of palm}(e) shows the relationship between the deformation performance and the air pressure of the palm Part A. It indicates that the splaying angle and force increase slowly when the applied air pressure increases from 10 to 60 kPa and then achieve a rapid increase afterward. Figure \ref{fig:experiment of palm}(a) shows the posture of the hand under 90 kPa air pressure with a 50\degree splaying angle. 
The air pressure should not be more than 100 kPa in order to avoid collapse.

The influence of the gravity on palm bending can be observed in Figure \ref{fig:experiment of palm}(b) and (c). The bending angles of the palm with down pose are obviously larger than those with the palm up pose. The tested maximum angle of the palm bending motion is up to 68\degree and of the thumb abduction is about 90\degree. The two bending ranges can meet most of the common grasping. Without the fibre-reinforcement, the two actuators of the palm part B should be driven under the safe range of air pressure less than 40 kPa. However, when the object is grasped and in contact with the hand, the pressure can be manually increased to enhance the reliability of the grasping.

\subsection{Grasping in real-world}
\begin{figure}[!hbp]
    \centering
    \includegraphics[width=\linewidth]{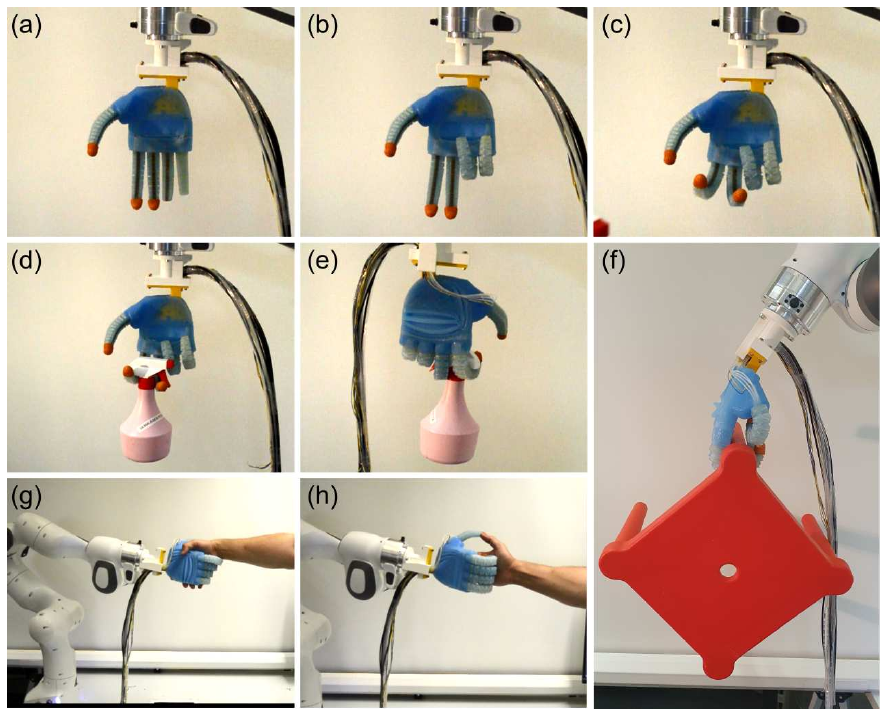}
    \caption{(a)-(e) present the capability of our soft hand prototype of performing lifting a watering can (143 g) by using 2 adjacent \acp{hbsf} (f) Grasping and lifting a chair (541 g) (g,h) Show the safely compliance in human-robot interaction.}
    \vspace{-.5cm}
    \label{fig:experiment of special application}
\end{figure}

\begin{figure*}[htbp]
    \centering
    \includegraphics[width=.9\linewidth]{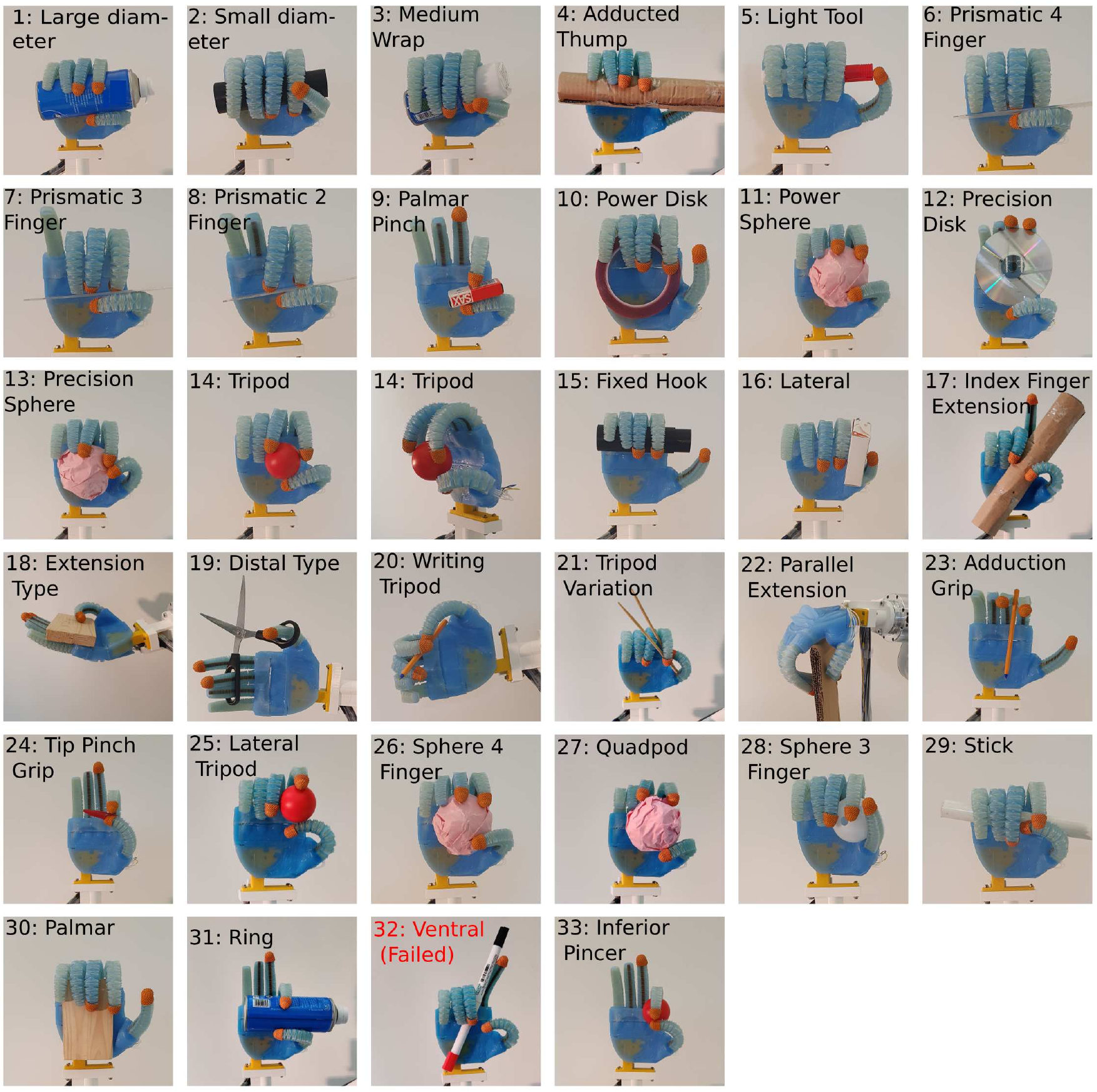}
    \caption{Enacted grasps of the Feix taxonomy. Grasps are numbered according to the Feix taxonomy \cite{Feix27}. Please check the attached video for the 32 grasps, \url{https://irobotics.aalto.fi/wp-content/uploads/2020/09/NewSoftHandvFast576p30.mp4} }
    \label{fig:experimets photos}
\end{figure*}

The deflection of the \acp{hbsf} can collaboratively work  used as a two-finger gripper to grasp and lift objects that allow the insertion of fingers into the neck portion, as shown in Figure \ref{fig:experiment of special application}(a)-(e). 
From  Figure \ref{fig:experiment of special application}(b) to (c), the index and middle \acp{hbsf} illustrate the deflection motion as illustrated in  Figure \ref{fig:designoffinger}(f). The gap between the index and middle fingers enlarges obviously. The bending motion can be used to provide the support force against gravity in case of the watering can. Besides, our hand is able to firmly grasp and lift the heavy and large-size chair in Figure \ref{fig:experiment of special application}(f). Figure \ref{fig:experiment of special application}(g) and (h) show the safety and compliance of the soft hand in human-robot interaction scenarios. 

\subsection{Grasp dexterity in Feix taxonomy}

To test the grasping performance of our hand, we implemented the grasping experiments according to the Feix taxonomy \cite{Feix27}, which includes 33 comprehensive grasp types. For every cases, the pressures are adjusted to reach the desired posture and the actuation sequences are generally follow the principle of actuating the palm first and then driving the fingers. The actuation of the soft  palm includes the splaying of the planar four fingers, palm bending and thumb abduction. It help to firstly fit the shape and size of the target objects and move the fingers in the proper position and orientation. Then the fingers are pressurized to implement the grasping tasks. The grasp quality is judged by moving up, down, and rotating the hand by Franka Emika Panda robotic arm under the speed of 40 mm/s. Furthermore, this procedure is repeated several times to evaluate the quality of the grasp.

Fig. \ref{fig:experimets photos} shows snapshots of 33 grasping posture types in Feix taxonomy \cite{Feix27}. Our hand failed to perform one grasping posture out of the 33. The ventral posture numbered as 32 failed because the object (marker pen used in 32) is thin and long, so the hand could not grasp it firmly. As the bending profile of the soft fingers has a circular shape, the inner diameter of the bending fingers is larger than the diameter of the marker pen, even by reaching a maximum bending angle.

\section{Conclusion}
In this paper, we have successfully developed a new soft humanoid hand capable of grasping different kinds of objects robustly. The hand exhibits the advantages of large grasping force, low cost, lightweight and potential applications to special cases. The pneumatic actuation enables the quick response of grasping in high compliance without damaging the objects. 
Meanwhile, we proposed a new design of soft finger, \ac{hbsf}, and a novel soft palm with 2 parts. The functions of each soft actuators were simulated  based on \ac{fem} in SOFA. The experimental results on the 6 fingers with different account of segments of the \ac{hbsf}  show that the soft finger with 11 segments is the best choice for achieving both the bending and deflection motions.

The main advantage of this study is that the postures of the hand can be adjusted with the help of the soft palm, and then
we can use different configurations to realise stable and dexterous grasping. The hand allows us to achieve 32 out of 33 grasp postures in Feix taxonomy, which demonstrates the postural dexterity of our soft hand.

A design limitation is that the thumb abduction angle and fingers splaying angle are limited by the silicone tensile strength and wall thickness of the air chamber structure. And the Flex sensors embedded in soft finger are not well used in the controller. Future work would address the optimisation of the soft palm to achieve that the thumb can touch to little finger and improve the control with safety and sensing.


%


\ifCLASSOPTIONcaptionsoff
  \newpage
\fi



%
\bibliographystyle{IEEEtran}
\bibliography{ref.bib}

\begin{thebibliography}{10}
\providecommand{\url}[1]{#1}
\csname url@samestyle\endcsname
\providecommand{\newblock}{\relax}
\providecommand{\bibinfo}[2]{#2}
\providecommand{\BIBentrySTDinterwordspacing}{\spaceskip=0pt\relax}
\providecommand{\BIBentryALTinterwordstretchfactor}{4}
\providecommand{\BIBentryALTinterwordspacing}{\spaceskip=\fontdimen2\font plus
\BIBentryALTinterwordstretchfactor\fontdimen3\font minus
  \fontdimen4\font\relax}
\providecommand{\BIBforeignlanguage}[2]{{%
\expandafter\ifx\csname l@#1\endcsname\relax
\typeout{** WARNING: IEEEtran.bst: No hyphenation pattern has been}%
\typeout{** loaded for the language `#1'. Using the pattern for}%
\typeout{** the default language instead.}%
\else
\language=\csname l@#1\endcsname
\fi
#2}}
\providecommand{\BIBdecl}{\relax}
\BIBdecl

\bibitem{Raphael35}
R.~Deimel and O.~Brock, ``A novel type of compliant and underactuated robotic
  hand for dexterous grasping,'' \emph{The International Journal of Robotics
  Research}, vol.~35, no. 1-3, pp. 161--185, 2016.

\bibitem{Xiaomin1}
X.~Liu, Y.~Zhao, D.~Geng, S.~Chen, X.~Tan, and C.~Cao, ``Soft humanoid hands
  with large grasping force enabled by flexible hybrid pneumatic actuators,''
  \emph{Soft Robotics}, ahead of print.

\bibitem{Shintake30}
\BIBentryALTinterwordspacing
J.~Shintake, V.~Cacucciolo, D.~Floreano, and H.~Shea, ``Soft robotic
  grippers,'' \emph{Advanced Materials}, vol.~30, no.~29, p. 1707035, 2018.
  [Online]. Available:
  \url{https://onlinelibrary.wiley.com/doi/abs/10.1002/adma.201707035}
\BIBentrySTDinterwordspacing

\bibitem{Mosadegh24}
B.~Mosadegh, P.~Polygerinos, C.~Keplinger, S.~Wennstedt, R.~F. Shepherd,
  U.~Gupta, J.~Shim, K.~Bertoldi, C.~J. Walsh, and G.~M. Whitesides,
  ``Pneumatic networks for soft robotics that actuate rapidly,'' \emph{Advanced
  Functional Materials}, vol.~24, no.~15, pp. 2163--2170, 2014.

\bibitem{6766586}
K.~C. {Galloway}, P.~{Polygerinos}, C.~J. {Walsh}, and R.~J. {Wood},
  ``Mechanically programmable bend radius for fiber-reinforced soft
  actuators,'' in \emph{2013 16th International Conference on Advanced Robotics
  (ICAR)}, 2013, pp. 1--6.

\bibitem{Lotfiani7}
A.~Lotfiani, H.~Zhao, Z.~Shao, and X.~Yi, ``{Torsional Stiffness Improvement of
  a Soft Pneumatic Finger Using Embedded Skeleton},'' \emph{Journal of
  Mechanisms and Robotics}, vol.~12, no.~1, 10 2019.

\bibitem{Abondance855}
S.~{Abondance}, C.~B. {Teeple}, and R.~J. {Wood}, ``A dexterous soft robotic
  hand for delicate in-hand manipulation,'' \emph{IEEE Robotics and Automation
  Letters}, vol.~5, no.~4, pp. 5502--5509, 2020.

\bibitem{Yilin97}
Y.~Sun, Q.~Zhang, and X.~Chen, ``Design and analysis of a flexible robotic hand
  with soft fingers and a changeable palm,'' \emph{Advanced Robotics}, ahead of
  print.

\bibitem{Clark65}
C.~B. Teeple, T.~N. Koutros, M.~A. Graule, and R.~J. Wood, ``Multi-segment soft
  robotic fingers enable robust precision grasping,'' \emph{The International
  Journal of Robotics Research}, ahead of print.

\bibitem{Polygerinos94}
P.~{Polygerinos}, Z.~{Wang}, J.~T.~B. {Overvelde}, K.~C. {Galloway}, R.~J.
  {Wood}, K.~{Bertoldi}, and C.~J. {Walsh}, ``Modeling of soft fiber-reinforced
  bending actuators,'' \emph{IEEE Transactions on Robotics}, vol.~31, no.~3,
  pp. 778--789, 2015.

\bibitem{Fras03}
J.~{Fras} and K.~{Althoefer}, ``Soft biomimetic prosthetic hand: Design,
  manufacturing and preliminary examination,'' in \emph{2018 IEEE/RSJ
  International Conference on Intelligent Robots and Systems (IROS)}, 2018, pp.
  6998--7003.

\bibitem{Mahmoud77}
M.~H. Mohamed, S.~H. Wagdy, M.~A. Atalla, A.~R. Youssef, and S.~A. Maged, ``A
  proposed soft pneumatic actuator control based on angle estimation from
  data-driven model,'' \emph{Proceedings of the Institution of Mechanical
  Engineers, Part H: Journal of Engineering in Medicine}, vol. 234, no.~6, pp.
  612--625, 2020.

\bibitem{Galloway19}
K.~C. Galloway, K.~P. Becker, B.~Phillips, J.~Kirby, S.~Licht, D.~Tchernov,
  R.~J. Wood, and D.~F. Gruber, ``Soft robotic grippers for biological sampling
  on deep reefs,'' \emph{Soft Robotics}, vol.~3, no.~1, pp. 23--33, 2016.

\bibitem{Weiping015}
W.~Hu and G.~Alici, ``Bioinspired three-dimensional-printed helical soft
  pneumatic actuators and their characterization,'' \emph{Soft Robotics},
  vol.~7, no.~3, pp. 267--282, 2020.

\bibitem{Feix27}
T.~{Feix}, J.~{Romero}, H.~{Schmiedmayer}, A.~M. {Dollar}, and D.~{Kragic},
  ``The grasp taxonomy of human grasp types,'' \emph{IEEE Transactions on
  Human-Machine Systems}, vol.~46, no.~1, pp. 66--77, 2016.

\bibitem{allard2007sofa}
J.~Allard, S.~Cotin, F.~Faure, P.-J. Bensoussan, F.~Poyer, C.~Duriez,
  H.~Delingette, and L.~Grisoni, ``Sofa-an open source framework for medical
  simulation,'' in \emph{Medicine Meets Virtual Reality (MMVR15)}, 2007.

\bibitem{Duriez138}
C.~{Duriez}, ``Control of elastic soft robots based on real-time finite element
  method,'' in \emph{2013 IEEE International Conference on Robotics and
  Automation}, 2013, pp. 3982--3987.

\bibitem{Coevoet362}
E.~Coevoet, T.~Morales-Bieze, F.~Largilliere, Z.~Zhang, M.~Thieffry,
  M.~Sanz-Lopez, B.~Carrez, D.~Marchal, O.~Goury, J.~Dequidt, and C.~Duriez,
  ``Software toolkit for modeling, simulation, and control of soft robots,''
  \emph{Advanced Robotics}, vol.~31, no.~22, pp. 1208--1224, 2017.

\end{thebibliography}

\end{document}